# A Mining Method to Create Knowledge Map by Analysing the Data Resource

Arti Gupta [1] , Prof. N. T. Deotale [2]

[1] *M.Tech scholar , Dept of Computer Science & Engineering, Rashtrasant Tukdoji Maharaj Nagpur University, Priyadarshini Bhagwati College Of Engineering , Harpur Nagar, Nagpur ,India*
[2] *Asst Prof , Dept of Computer Science & Engineering, Rashtrasant Tukdoji Maharaj Nagpur University, Priyadarshini Bhagwati College Of Engineering, Harpur Nagar, Nagpur ,India*

*Abstract*—**The fundamental step in measuring the robustness of a system is the synthesis of the so-called Process Map. This is generally based on the user's raw data material. Process Maps are of fundamental importance towards the understanding of the nature of a system in that they indicate which variables are causally related and which are particularly important. This paper represent the system Map or business structure map to understand business criteria studying the various aspects of the company. The business structure map or knowledge map or Process map are used to increase the growth of the company by giving some useful measures according to the business criteria. This paper also deals with the different company strategy to reduce the risk factors. Process Map is helpful for building such knowledge successfully. Making decisions from such map in a highly complex situation requires more knowledge and resources.**

## I INTRODUCTION

Over the next few years there should be an increased emphasis on taxonomies, ontologies and Knowledge Management tools. These are very much helpful in the method of creation the commonly-shared knowledge bases. Knowledge Map as a one of the Knowledge representation tools is a navigation aid to expore information and useful knowledge, showing the importance and the relationships between knowledge stores. This generic tool was specifically made for Knowledge Management. It is a proper to remember that Knowledge Map does not store knowledge. It just points to people who own it.

Many companies around the world already use Knowledge Maps to transform their employees into knowledge partners, focusing the organization on the most critical issues facing the business first step of creation useful Knowledge Management system in organization. These maps allow employees which are responsible for Knowledge Management to built the KM system based on their own document.

Knowledge Management (KM) refers to a multi-disciplined approach to achieving organizational objectives by making the best use of knowledge. KM focuses on processes such as acquiring, creating and sharing knowledge and the cultural and technical foundations that support them

The following diagram describe the main technologies that currently support knowledge management systems.

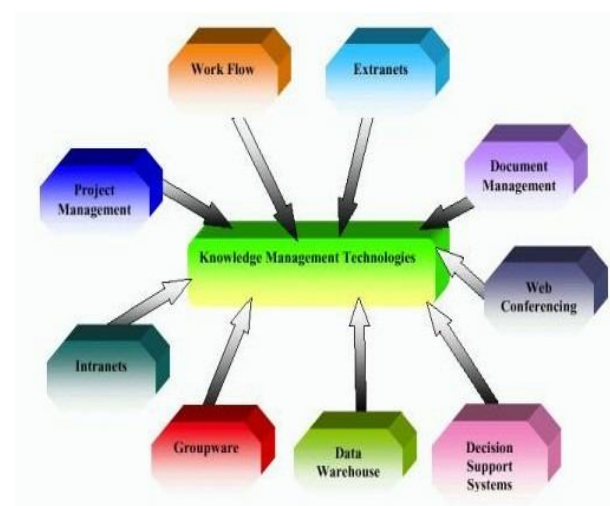

Fig.1 Knowledge management technologies

These technologies is correlated with to four main stages of the KM life cycle:

■ Knowledge is gained or captured using intranets, extranets, groupware, web conferencing, and document management systems.

■ An organizational memory is generated by refining, organizing, and storing knowledge using structured repositories such as data warehouses.

■ Knowledge is shared through education, training programs, automated knowledge based systems, expert networks etc.

■ Knowledge is applied or leveraged for further learning and innovation via mining of the organizational memory and the application of expert systems such as decision support systems.

All of these stages are enhanced by effective workflow and project management.

The Implications of Knowledge Management are for:-

• *Database End User :* From business class users to the general public, database users will enjoy a new level of interaction with the Knowledge Map  system including just-in-time knowledge that delivers precise relevant information on demand and in context. The more complicate , smart systems will translate to optimal usability and less time spent searching for relevant information.   For example, data analysts will enjoy simplified access and more powerful tools for data exploitation. The use of knowledge bases can reduce customer service costs by providing customers with easy access to 24/7 self service through  smart systems that reduce the need to contact customer service or technical support staff.  Database users may even create customized views of knowledge bases that support their needs.

• *Database Core Developers*: The design and development of knowledge based systems will be considerably more critical   than current database development methods.  Developers must consider the overall technical structure of the corporation to ensure seamless interoperability.  The use of standardized meta data and methods will also facilitate both intra-corporate and inter-corporate interoperability.   Making effective physical storage and platform choices will be equally more complex.  Both knowledge base developers and administrators must understand the role of the knowledge base in the overall Knowledge Map system.

• *Database Administrators or DBA*: Database Administrators or DBA will evolve into Knowledge Managers.  The knowledge base will store and maintain corporate   memory and Knowledge Managers will become the gatekeepers of corporate knowledge. The lines between technical roles such as Web Developer, Data Analyst or Systems Administrator will blur as these systems merge into and overlap with KM systems. DBAs will need to have some knowledge about each of these disciplines.

## II PROBLEM DEFINATION

• *Fuzzy Cognitive Maps (FCM)*
A FCM is a graphical representation consisting of nodes indicating the most relevant factors of a decision support environment, and links between these nodes representing the relationships between those factors. FCM is a modelling methodology for complex decision making systems, which has originated from the combination of fuzzy logic and neural networks. A FCM describes the nature  of a system in terms of concepts; each concept representing an entity, a state, a variable, or a characteristic of the system.

In the following fig no 2 , the positive edge rule says  that a survival threat increase run away. It is a positive causal connection. The runaway response either grows or falls as the threat grows or falls.

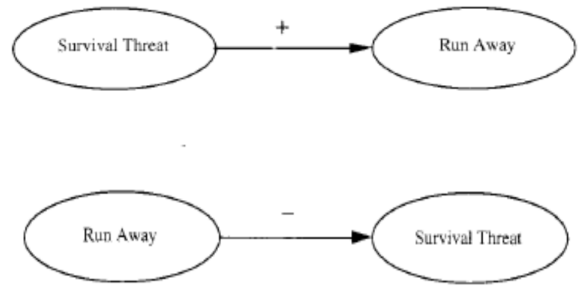

Fig 2  A Simple FCM example

The negative edge rule clears that running away from a predator lower the survival threat .It is a negative causal connection. The survival threat grows the less the pray runs away and falls the more the pray run away.

In FCM as  number of nodes  increases ,number of arcs also increased. So it becomes complicate  to understand. And  the  process  is  time  consuming.  To solve this problem, Allen and  Marczyk introduced the Knowledge Map concept  which  was  derived  from  FCM.  The Knowledge Map  is very much useful in  extracting knowledge  and also  shows  the flow of relationship between various datasets.

The  main  drawback  of FCM was  when  too  many parameters came into map, it become very complex to judge. So KM replaced this drawback by replacing this complex view by segmented view. The segmented view is selected because it reduces redundancy. The segmented theory is based on Design Structure Matrix (DSM) theory.

This  process  reduces   complex data  to a "Knowledge Map."  .and how  to  generate  Knowledge Maps from complex data sets and how Knowledge Maps can be used to help make better decisions.

## III RELATED WORK

A number of algorithms for learning FCM model structure have  been  introduced.  In  general  two  main  learning paradigms  are  used.  Hebbian  learning  and  genetic algorithms.  Dickerson  and  Kosko  proposed  simple Differential   Hebbian  Learning  law  (DHL)  to be applied to learning FCMs .The learning process iteratively updates values of weights of all edges from the FCM graph until the desired structure is found.

In 2002 Vazquez presented an extension to DHL algorithm by introducing new rules to update edge values . This new algorithm was called Balanced Differential Algorithm (BDA). The new algorithm eliminates the limitation of DHL method where weight update for an edge connecting two concepts (nodes) is dependent only on the values of these two concepts. Therefore proposed learning method was applied only to FCM with binary concept values which significantly restricts its application areas.

Another method based on Hebbian learning was proposed in 2003. Papageorgiou et al. developed an algorithm called Nonlinear Hebbian Learning (NHL) to learn structure of FCM . The main idea  is to update weights associated only with edges that are already suggested by expert, non-zero weights Therefore the NHL procedure allows obtaining model that retains structure which is enforced by the expert but at the same it requires human intervention before the learning process starts.

Active Hebbian Algorithm (AHL) introduced by Papageorgiu et al. in 2004 is the next attempt to develop Fuzzy Cognitive Map development. This approach introduces and exploits the task of determination of the sequence of activation concepts. The main limitation of this algorithm is that it still requires human interference.

In 2001 Koulouriotis et al. applied the Genetic Strategy (GS) to learn fuzzy cognitive map model structure weights of relationships from data.  In this procedure, the learning process is based on a collection of input/output pairs which are called examples Its main drawback is the need for multiple state vector sequences which might be difficult to obtain for many real-life problems.

Parsopoulos et al. in 2003 applied Particle Swarm Optimization (PSO) method which belongs to the class of Swarm Intelligence algorithms to learn FCM structure based on a historical data consisting of a sequence of state vectors that leads to a desired fixed-point attractor state. PSO is a population based algorithm which goal is to perform a search by maintaining and transforming a population of individuals. This method improves the quality of resulting fuzzy cognitive map  model by minimizing an objective function.

Next state of the art learning method for Fuzzy Cognitive Map, introduced by Stach et al. in 2005 applies real-coded genetic algorithm (RCG.A) to develop FCM model from a set of historical data .This approach is very flexible in terms of input data:\it can use either one time series or multiple sets of concepts values over successive iterations.

## IV  LITERATURE SURVEY

Author B. Kosko   developed a  fuzzy causal algebra for governing causal propagation on FCM.. FCM matrix representation and matrix operations are presented in this context. Dickerson and Kosko proposed a simple Differential Hebbian Learning (DHL) algorithm which iteratively updated the values of the weights until they converged to certain predefined state. The generated FCM then become very complicated and difficult to understand.

J. Aguilar proposed the automated construction of FCM using learning procedure is a new field. These approaches have lots of  advantages of quantified results but have several drawbacks. First thing is that the model typically requires a great deal of effort and specialized knowledge outside the domain of interest. And the second thing is that  systems involving significant feedback propagates casual influences in complicated chains may be nonlinear in which case a quantitative model may not be possible. FAs a result  numerical data may be hard to come.

J. L. Salmeron suggest to build an Augmented Fuzzy Cognitive Map based for modelling Critical Success Factors in Learning Management Systems. The study of Critical Success Factors helps decision makers to extract knowledge from the multidimensional learning process the core activities that are necessary for success.

## V  PROPOSED WORK

*A Module 1 : Requirement Gathering*

Collect different dataset from various companies and need to create the database to mine knowledge maps from it. These databases can be from financial companies or any companies having major finance background. And need to collect revenue, income tax returns and maintenance cost like parameter The database used is financial related, the charted account firm or Market share data of any financial firm. If any dataset is available for research we will use that, or we need to create our own  Knowledge map always represent the concepts. Concept can be anything related to your domain like profit, loss, expenses, customer satisfaction, hr satisfaction etc. So we are keeping the data related to these sections.

*\*.About UI & Software Used:*

UI is created in PHP- MySQL as it has quality graph and chart support in the form of P-Chart Library.  Our paper is mostly based on graph we choose PHP.

*\* Dataset Preparation*:

Most of companies refuse to give their confidential data even for student research purpose we need to create sample financial dataset for research based on real time values.

### B. Module 2 : Knowledge Map Creation

This is very important step of research as we are creating knowledge maps based on the historical data collected in module1. Knowledge map creation step has 4 steps

- Scatter plot generation
- Fuzzy Rule Generation
- Map construction
- Hubs & Inactive node identification

### 1)  Scatter Plot Generation:

A scatter diagram is a tool for analyzing relationships between two variables. One variable is plotted on the horizontal axis and the other is plotted on the vertical axis. The pattern of their intersecting points can graphically show relationship patterns. Most often a scatter diagram is used to prove or disprove cause-and-effect relationships. While the diagram shows relationships, it does not by itself prove that one variable *causes* the other. In addition to showing possible cause and-effect relationships, a scatter diagram can show that two variables are from a common cause that is unknown or that one variable can be used as a surrogate for the other.

Scatter diagrams will generally show one of  six possible correlations between the variables:

• *Strong Positive Correlation*
   The value of Y clearly increases as the value of  X increases.

• *Strong Negative Correlation*
   The value of Y clearly decreases as the value of  X increases.

• *Weak Positive Correlation*
   The value of  Y increases slightly as the value  of X increases.

• *Weak Negative Correlation*
   The value of  Y decreases slightly as  the  value of  X increases.

• *Complex Correlation*
   The value of Y seems to be related to the value of  X , but the relationship is not  easily determined.

• *No Correlation*
   There is no demonstrated connection between  the two variables

In order to generate scatter plots we first create the excel file that consist financial data upto 10 years i.e from year 2004 to year 2013.After creating the excel file, we have to first filter that cells so that it occupy less space. At the last we have to import that file into our database. After importing , it automatically takes the file.

The next process is to create the scatter plots from the available data of the file. If we want to plot a graph between Income v/s Expenses then we have to take "Income" on X axis and "Expenses" on Y axis. The "Income" consist parameters like Sales turnover, Excise duty ,Net Sales, Stock adjustments and "Expenses" consist Raw materials, Power and fuel cost, Employee cost, Other expenses etc. The scatter plot between Income versus Expenses looks like,

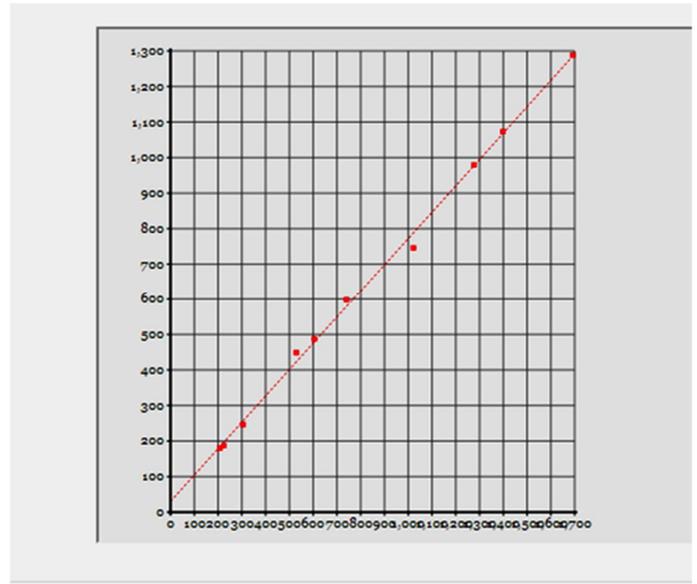

Fig. 3 Scatter plot of Income v/s Expenses

As in fig 2, the scatter plot shows a straight line ,it shows that company's loss and profit balance sheet is perfect. That means ,from year 2004 to 2013 the profit is continuously increases and in the same way growth of the company is also increases.

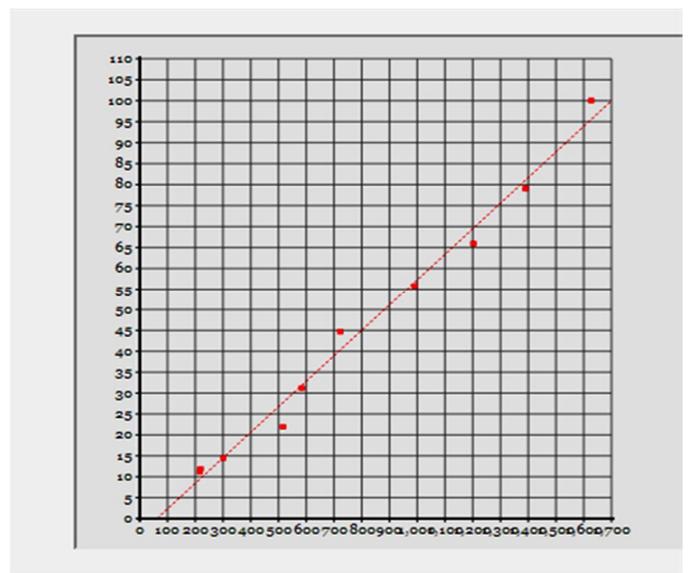

Fig. 4. Scatter plot of  Net Sales v/s Employee cost

Similarly fig 4 shows the relationship between Net Sales and Employee cost. Employee cost is nothing but the employee salary. There are few fluctuations between them as we seen it from the diagram. Whereas Net sales is on the X axis and Employee cost is on the Y axis .As the Net sales increases employee cost is also increases.

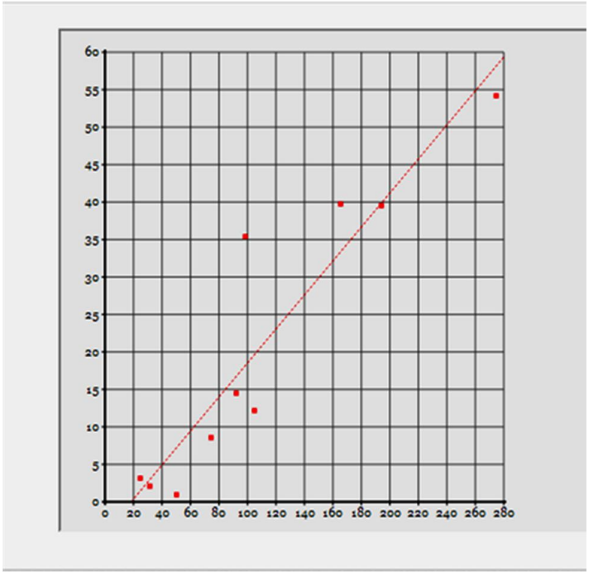

Fig. 5 Scatter plot of Profit before tax v/s Tax

Likewise in fig.5 the scatter plot shows the relationship between profit before tax v/s tax. The Profit before tax is on the X axis and Tax is on the Y axis. From that we can conclude that the profit before tax is not so as good after paying the tax. The tax is more than the profit.

2) *Fuzzy Rule Generation:*

Fuzzy set A on a universe of discourse U is characterized by a membership function that takes values in the interval [0, 1].

In 1975 Professor Ebrahim Mamdani of London University built one of the first fuzzy systems to control a steam engine and boiler combination. He applied a set of fuzzy rules supplied by experienced
human operators.

For ex

O  Input x: research _funding

O  Input y: project _staffing

O  Output z: risk

• *Rules*

Rule1: If research _funding is **adequate** or
    Project _staffing is **small** Then risk is **low.**
Rule2: If research _funding is **marginal** and

Project _staffing is **large** Then risk is **normal.**

Rule3: If research _funding is **inadequate** Then risk is **high.**

3) *Map Construction*

BSMs are of fundamental importance towards the understanding of the structure of a business. BSMs indicate which business parameters are related and which are particularly important.

In a Business Structure Map the following components are present

o  *Nodes or Concepts* - These are the business parameters located along the map's diagonal and represented by red squares or discs. The node represent the concept or entity.

Examples: profit, loss, expenses, customer satisfaction, hr satisfaction etc can be represented as node.

There are two type of nodes that is input and output. Input nodes are those which brings input to the firm and output goes are output nodes.

Example: Revenue is input node and taxes are output node Input and output nodes are represented by 2 different colours.

o  *Links or Connectors* – These are either strong or weak and are represented by black and grey connectors respectively. These links represent the strong and weak dependency between various parameters. These dependencies can be represented by the different colour connectors.

o  *Hubs* – The red discs correspond to those parameters which are related to the highest number of other parameters.

Online self-rating system delivers the so-called Business Structure Map which reflects the relationships between the various parameters of your business. The full power of our technology lies in its ability to analyze the complexity of systems of systems. Think of portfolios of financial products, companies or conglomerates of corporations and banks. How are they related to each other , How do they interact, Which are the most important ones, Which would cause the largest amount of damage in the event of default ,How would contagion propagate, All these and other questions may be answered with our tools in a truly innovative manner.

4) *Hubs and inactive nodes identification:*
The number of links in a node decide which node is hub and which node is inactive node. The node with the most relationships is called the hub, while node with no links are inactive. Both kinds of nodes are clearly represented in the map. The hubs that have close relationships with other nodes appear as circular shapes while the inactive nodes are shown as white squares.

*C. Module 3: Analysis*

*1) Static Analysis:* The static analysis uses all the historical data to construct a KM to give complete picture of companies.

*2) Time Domain Analysis*: In time domain analysis the data will be split into several continues periods or windows. For ex ,to view the salary of particular month or year.

## VI RESULTS

Knowledge map is useful for representing knowledge and for monitoring the health of companies. Furthermore sudden changes of the key features of the K.M should be taken seriously by policymakers as an alarm of a crisis. The main objective is to get good results from historical data so that the prediction and formation of policies. This method can be used in many fields, such as product design, management, medicine and air traffic control. maps mined from historical data are more valid and lose less information than those relying on the perceptions of experts..The K.M mining method proposed by Marczyk eliminates the long iterative procedure and constructs a K.M directly by analyzing the data resource.

Based on knowledge map you can judge which parameters has the greatest impact on the company policy. You can see the relationship between employee salary and productivity.

(1) Identification of cause-effect relationships between variables,

(2) Visualization of how information flows within a given system, and

(3) Ranking of variable importance indicated by hubs and inactive nodes

When we will construct knowledge map with regular parameter, we will add few more parameters based on the output of KM and which factors are affecting lot.

## VII CONCLUSION

This paper describes a mining method to construct knowledge maps utilizing historical data without the intervention of domain experts. The software is used to apply the method to analyze component stock corporations. The static analysis results show that the KM is capable of discovering the structure of the examined systems given through fuzzy rules. Time-domain analysis reveals the evolution of the main feature so the knowledge maps which can be used by policymakers to monitor the company health. The results demonstrate that the mining of knowledge main properties of the KM can effectively indicate crises, which is not possible by conventional risk rating methods.

The scatter plot is the basis of the knowledge map. It helps to understand the relationship between various parameters of the company

This idea is very much useful in automobile and aerospace industries as a design simulation. It will be also very helpful to other related areas such as textile industries, bank , sales etc. Banking, E-commerce, HR, and Production industries.

## ACKNOWLEDGEMENT

The concepts presented represent the work of Dr. Jacek Marczyk which is greatly acknowledge.The information in this document reflects solely the views of the authors.

## REFERENCES

[1]   B. Kosko, Fuzzy cognitive maps, International  Journal of Man-Machin  Studies, vol. 24, no. 3,  pp.

[2]   J. Aguilar, A survey about fuzzy cognitive map  papers, Internationa Journal of Computational  Cognition, vol. 3,no. 2, pp. 27-33, 2005.

[3 ]   L. Rodriguez-Repiso, R. Setchi , and J. L. Salmeron, Modelling IT projects success with  fuzzy cognitive maps,Expert Systems with Applications, vol. 32, no. 2, pp. 543-559, 2007.

[4]   Z. Peng, B. Yang, C. Liu , Z. Tang, and J. Yang, Research on one fuzzy cognitive map  classifier, (in Chinese), Application Research of Computers , vol. 26 , no. 5 , pp.1757-1759,  2009.

[5]   T. Hong and I. Han , Knowledge-based data mining of news information on the Internet  using cognitive maps and neural networks, Expert Systems with Applications, vol. 23,  no. 1, pp. 1-8, 2002.

[6]   E. I. Papageorgiou , Learning algorithms for  Fuzzy cognitive mapsla  review study , IEEE  Trans on Systems, Man and Cybernetics , vol. 42, no. 2, pp. 150-163, 2012.

[7]   J. A.Dickerson and B. Kosko,Virtual worlds as  fuzzy cognitive maps  Presence , vol. 3 , no. 2, pp. 173-189, 1994.

[8]   M. Schneider, E. Shnaider , A. Kande l, and G. Chew , Constructing fuzzy cognitive maps, in  Proc. 1995 IEEE International Conference on  uzzy Systems , Yokohama , Japan , 1995, pp.  2281-2288.

[9]   K. E. Parsopoulos, E. I. Papageorgiou , P. P. Groumpos, and M. N. Vrahatis, A first study of  Fuzzy cognitive maps learning using  Particle swarm optimization, in Proc. 2003 Congress on Evolutionary Computation, 2003, pp. 1440  1447.

[10]   W. Stach, L. Kurgan , W. Pedrycz, and M. Reformat , Learning Fuzzy   cognitive maps with required precision using genetic algorithm  approach , Electronics Letters, vol.40, no. 24, pp. 1519-1520, 2004.

[11]   G. Allen and J. Marczyk ,Tutorial on  Complexity management for decision-making,  pdf, 2012.

[12]   J. Marczyk ,A New Theory of Risk And Rating, Trento:Editrice Uni Service , 2009.

[13]   D. V. Stewar,The design structure matrix: A Method for managing  the design of complex  systems, IEEE Transactions on Engineering Management,vol. EM-28, no. 3, pp.71-74,1981.

[14]   S. Aumonier , Generalized correlation power  Analysis , in Proc. ECRYPT  Workshop on  Tools For Cryptanalysis Krakw,Poland,2007.

[15]   C. E. Shannon , A mathematical theory of  communication, Bell System Technical Journal, vol. 27, pp.379-423, 1948.

[16]   B. Lent, A. Swami , and J. Widom , Clustering  Association rules, in Proc. 13th International Conference on DataEngineering,Birmingham,  England, 1997, pp. 220